\documentclass[runningheads]{llncs}
\usepackage{graphicx}
\usepackage{csquotes}
\usepackage{float}
\usepackage{amssymb}
 
\begin{document}

\title{Enhanced Behavioral Cloning Based self-driving Car Using Transfer Learning}
\subtitle{}

\titlerunning{Enhanced Transfer Learning Model For self-driving}        % if too long for running head
\author{Uppala Sumanth\inst{1}\orcidID{0000-0002-6412-5123} \and
Narinder Singh Punn\inst{1}\orcidID{0000-0003-1175-1865}
\and
Sanjay Kumar Sonbhadra\inst{1}\orcidID{0000-0002-7457-9655}
\and
Sonali Agarwal\inst{1}\orcidID{0000-0001-9083-5033} 
%Third Author\inst{3}\orcidID{2222--3333-4444-5555}
}
\authorrunning{Sumanth et al.}

% \authorrunning{} % if too long for running head
\institute{ Indian Institute of Information Technology Allahabad, Jhalwa, Prayagraj, Uttar Pradesh, India \\
\email{\{mit2018075,pse2017002,rsi2017502,sonali\}@iiita.ac.in}}
\maketitle              % typeset the header of t

\begin{abstract}
With the growing phase of artificial intelligence and autonomous learning, the self-driving car is one of the promising area of research and emerging as a center of focus for automobile industries. Behavioral cloning is the process of replicating human behavior via visuomotor policies by means of machine learning algorithms. In recent years, several deep learning-based behavioral cloning approaches have been developed in the context of self-driving cars specifically based on the concept of transfer learning. Concerning the same, the present paper proposes a transfer learning approach using VGG16 architecture, which is fine tuned by retraining the last block while keeping other blocks as non-trainable. The performance of proposed architecture is further compared with existing NVIDIA’s architecture and its pruned variants (pruned by 22.2\% and 33.85\% using $1\times1$ filter to decrease the total number of parameters). Experimental results show that the VGG16 with transfer learning architecture has outperformed other discussed approaches with faster convergence.

\end{abstract}

\section{Introduction}
\label{intro}
\begin{figure}[H]
\centering
\includegraphics[scale=0.25]{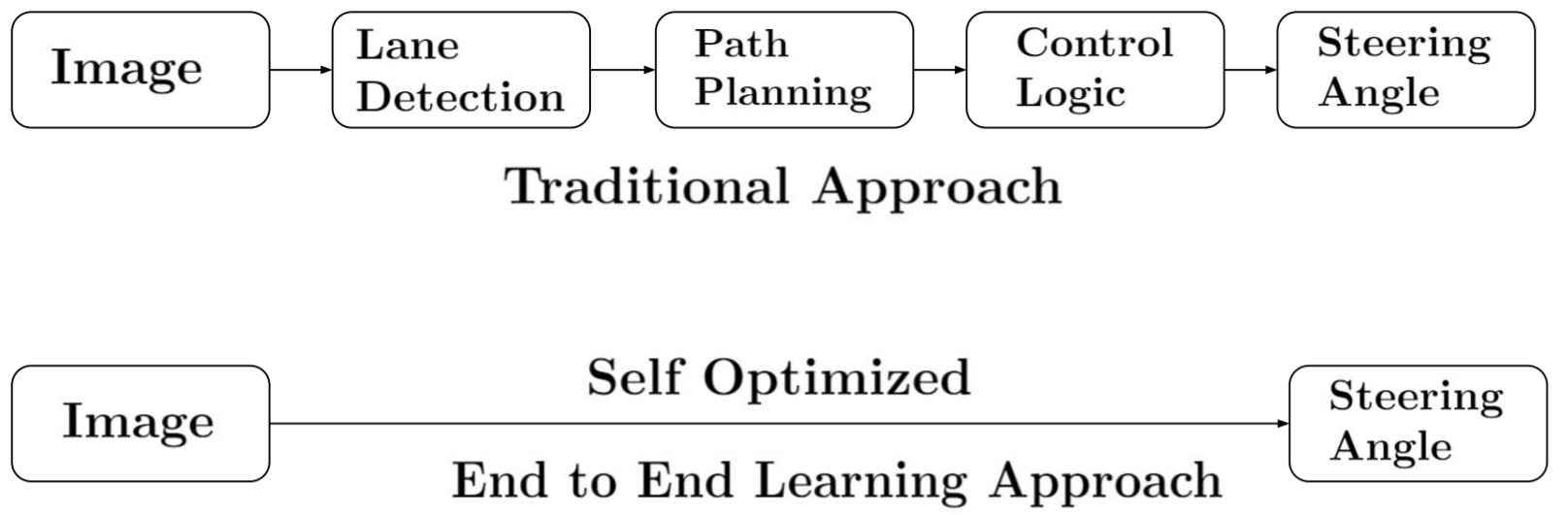}
\caption{End-to-end learning.}
\label{fig1}       
\end{figure}

The end-to-end deep learning model is the most popular choice to deal with large volume data~\cite{chopra2020end,lee2020autonomous,chen2017end,glasmachers2017limits} among researchers. Conventionally, the deep learning approaches decompose the problem in several subproblems to solve them independently, and all the outputs are combined to draw final decision. Many automobile companies like Hyundai, Tesla, etc., are trying to bring millions of self-driving or autonomous cars on the road by utilizing deep learning approaches. In this frantic race to come up with fully safe self-driving cars, some of the organizations like NVIDIA is following the end-to-end approach~\cite{bojarski2017explaining} as shown in Fig.~\ref{fig1}, whereas Google is following mid-to-mid approach~\cite{bansal2018chauffeurnet}. Following from these notions, the main objective of present research work is to predict the steering angle of the car via front facing camera.

The behavioral cloning~\cite{WEBSITE:6} is a process of reproducing human performed tasks by a deep neural network. behavioral cloning is achieved by training the neural network with the data of human subject performing the task. In 1989, a self-driving car was developed by Pomerleau~\cite{pomerleau1989alvinn} based on neural networks. Afterwards, since past 130 years, the automobile manufacturers did not give attention to the replacement of the driver, who is the most vulnerable part of the car. The automotive companies tried to make the cars safer by including many safety features like anti-lock braking systems, shatter-resistant glass, airbags, etc~\cite{wiki:safety}. However, organizations failed to succeed in developing driver-less intelligence. Self-driving cars are the most desirable revolutionary change in 21st century for a fully safe driving experience to change the way of transportation. According to the World Health Organization’s report on \enquote{Global status report on road safety 2018}, every year around 1.35 million humans lose their lives due to road accidents~\cite{world2018global}. Self-driving cars will bring this number down and also enable people with disabilities to commute easily. Convolution neural networks (CNN) have revolutionized Pattern recognition~\cite{lecun1989backpropagation} with the ability to capture 2D images in the context of self-driving cars. The greatest advantage of CNN is that it automatically extracts the important features to interpret the surrounding environment from the images which can be utilized to develop the intelligent driving system. 

In the present research, to establish the importance of transfer learning approach concerning self-driving cars, a novel end-to-end based VGG16 approach is proposed which is fine-tuned to predict the steering angle based on the environmental constraints. Later, the proposed approach is compared with NVIDIA and its pruned variants. Due to the lesser number of parameters in the pruned architectures, the training time reduces significantly compared to baseline architecture. Since the transfer learning approach follows from the pre-trained model where only a part of network is trained, significant computational time is saved without compromising the performance. It has been observed that if the tasks are similar then the weights of initial few layers are similar and the last layers  have relevant information towards the task~\cite{yosinski2014transferable}; making transfer learning a better way of saving training time. 

The paper is organised in various sections including related work which briefly discusses the available approaches applied to self-driving cars and highlights the drawbacks and advantages of the research work carried out so far. The proposed approach section presents the various approaches utilized in the process of generating a novel model which accurately drives the car. Dataset and pre-processing techniques are also discussed in the subsequent sections. At the end, the experimental results were elaborated with concluding remarks. 
 
\section{Related Work}
The process of reconstructing the human subcognitive skill through the computer program is referred as behavioral cloning. Here, actions of human performing the skill are recorded along with the situation that gave rise to the action~\cite{torabi2018behavioral}. There are two popular methods for behavioral cloning. In the first method, the skill could be learned in terms of series of dialogues with an operator. Here, in case of autonomous vehicle, it is anticipated from the operator to explain all set of useful skills to control the vehicle. This method is challenging because manual description of skills is not perfectly possible due to human limitations. Alternatively, skill can be reconstructed through recorded actions which are maintained in a structured way by using learning algorithms in terms of various manifestation traces~\cite{sammut2011encyclopedia,michie1994building,kulic2006autonomous,michie1993knowledge} to reproduce the skilled behavior.

\begin{figure}
\centering
\includegraphics[scale=0.35]{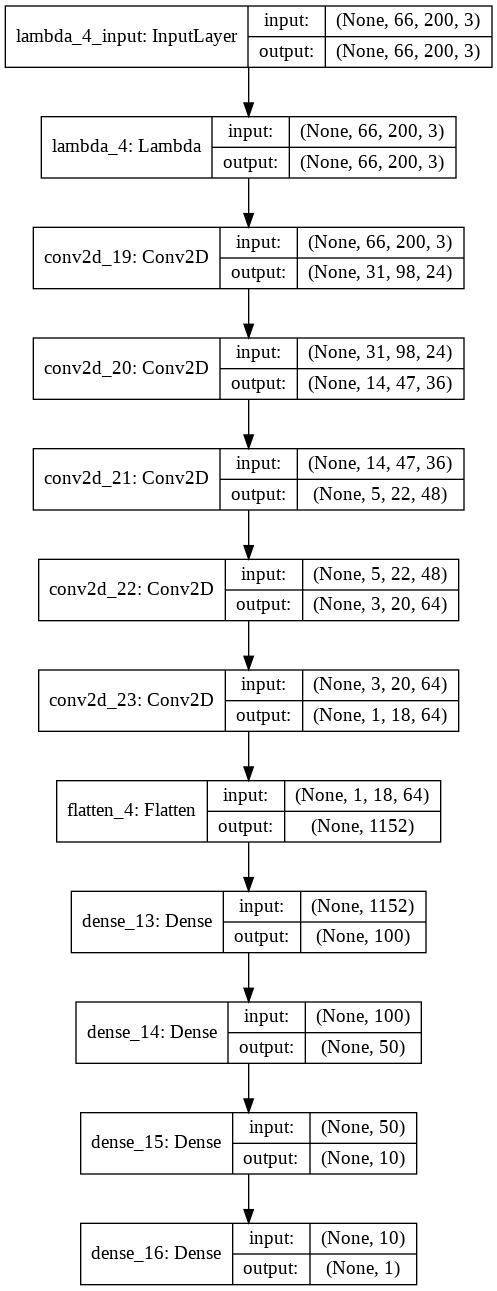}
\caption{Baseline NVIDIA architecture.}
\label{fig2}      
\end{figure}

Defense Advanced Research Projects Agency (DARPA) initiated DARPA autonomous vehicle(DAVE)~\cite{lecun2004dave} project including a radio-controlled model truck which is attached with sensors and lightweight video cameras to test vehicle in an intrinsic environment having trees, heavy stones, lakes, etc. The testing vehicle is trained with 225000 frames of driving data. However, in test runs, DAVE crashed for every 20 metres on an average. In 1989 Pomerleau built an autonomous land vehicle in a neural network (ALVINN) model using the end-to-end learning methodology and it was observed that the vehicle can be steered by a deep neural network~\cite{pomerleau1989alvinn}. NVIDIA started their research work in the self-driving inspired by the ALVINN and DARPA projects. The motivation for their work was to create an end-to-end model which enables steering of the car without manual intervention~\cite{bojarski2016end,bojarski2017explaining} while recording humans driving behavior along with the steering angle at every second by using controller area network (CAN) bus. Based on the NVIDIA proposed architecture pilotnet (as shown in Fig.~\ref{fig2}), Texas Instruments released JacintoNet i.e an end-to-end neural network for embedded system vehicles such as tiny humanoid robots~\cite{viswanath2018end}. In 2020, a group of researchers from Rutgers university proposed a feudal network based on hierarchical reinforcement learning that performs similar to the state-of-the-art models with simpler and smaller architecture which reduces training time~\cite{johnson2020feudal}. Jelena et al.~\cite{kocic2019end} have proposed a network which is 4 times smaller than the NVIDIA model and about 250 times smaller than the AlexNet~\cite{krizhevsky2012imagenet}. The model is developed only for the use in the embedded automotive platforms. To study the working of these end-to-end models Kim et al.~\cite{kim2017interpretable} researched about the region of the images contributing in predicting steering angle.  

Although learning to drive from this system would not suffice the self-driving car, furthermore the driving system should also address the issues such as how it would backtrack on to the road if it goes off the road by mistake, or else the vehicle will eventually move out of the road. Therefore, the images which are provided by the dataset, are combined with more images to visualize the vehicle in different field-of-views on and off the road. The datasets usually augmented with new images generated by view transformations via flipping the images to cover maximum possible scenarios~\cite{bojarski2017explaining}. For the transformed images the steering angle is changed in such a way that the vehicle would come back to the right position and direction within 2 seconds. The NVIDIA model proved to be quite powerful by achieving 98\% autonomy time on road. The results observed from NVIDIA's model consisting of only 5 convolution layers followed by 3 dense layers exhibited limited performance, thus it is evident that complex tasks require complex structure of deep neural networks with more number of layers to achieve better performance.

\section{Proposed approach}
A competent human is required for controlling any intricate system such as helicopter, bike, etc. The competency is learnt through experience that develops within the subconscious capability of the brain. These subcognitive skills are challenging and can only be described roughly and inconsistently. In case of frequently occurring actions, the competency can be achieved by the system via learning from the recorded common patterns using deep learning techniques. Extracting and replicating such patterns from human subject performing the task is called behavioral cloning~\cite{bain1995framework}.

\begin{figure}
\centering
\includegraphics[width= 5.5cm, height=13cm]{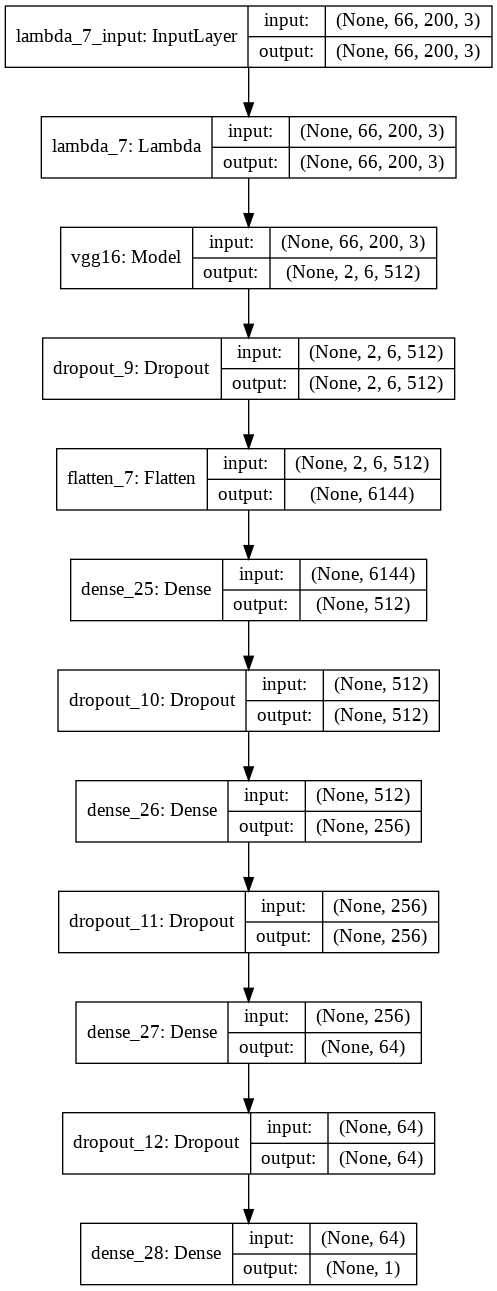}
\caption{Architecture of VGG16 With Transfer Learning.}
\label{fig03}     
\end{figure}

\begin{figure} 
\centering
\includegraphics[scale=0.15]{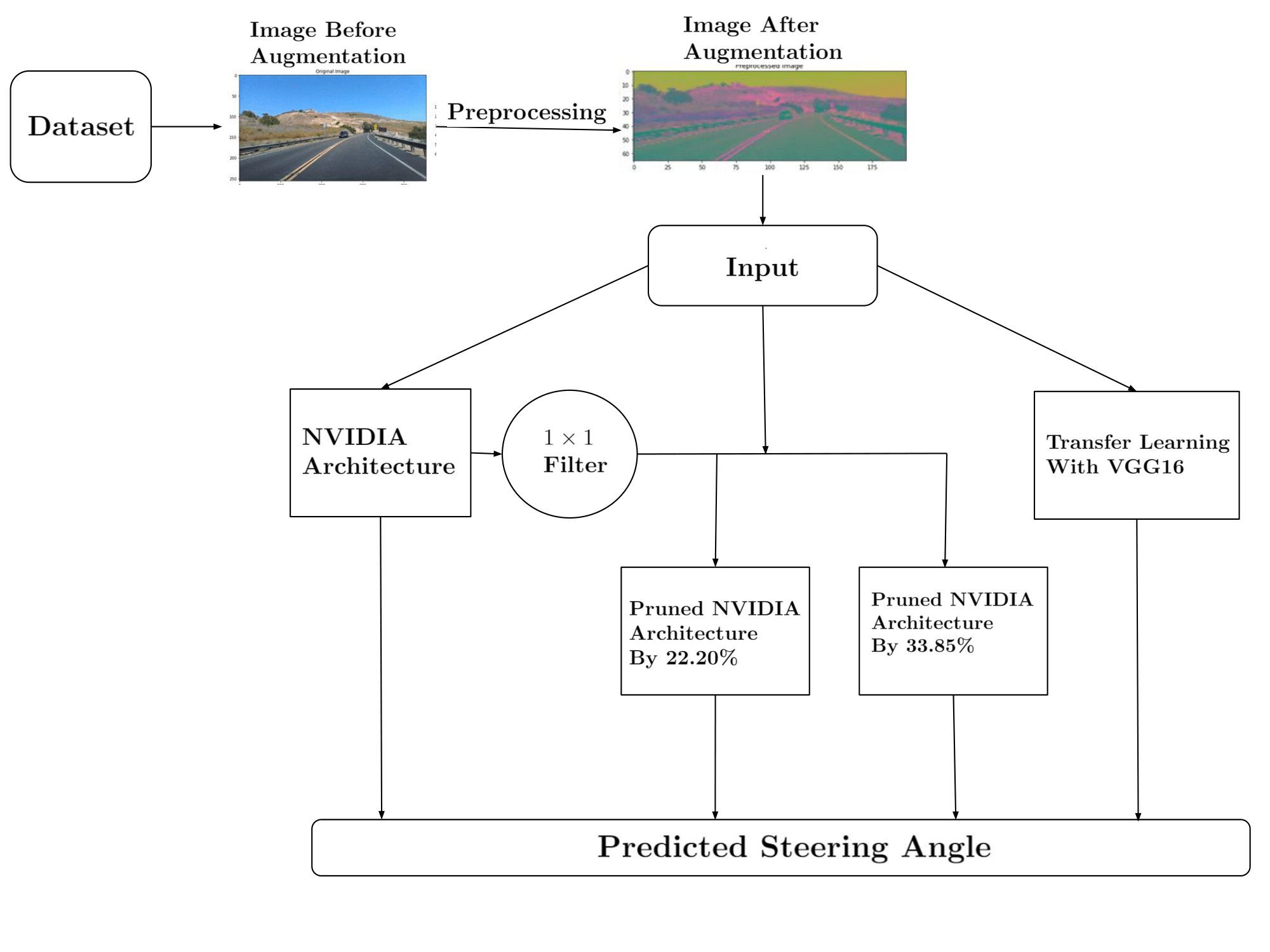}
\caption{Schematic representation of proposed approach.}
\label{fig3}       
\end{figure}

Following from the idea of behavioral cloning, a novel end-to-end transfer learning based VGG16 approach (as shown in Fig.~\ref{fig03}) is proposed to predict the appropriate steering angle. The proposed model is compared with NVIDIA baseline model and its pruned variants built by chopping off the baseline NVIDIA model by 22.2\% and 33.85\% by using a $1\times1$ convolution filter. Fig.~\ref{fig3} presents the overall schematic representation of the proposed approach.

\subsection{Network pruning using $1\times1$ filter}
\subsubsection{The use of $1\times1$ convolution for network pruning}
To downsample the contents of feature maps pooling is used which will reduce the height and weight whilst retaining the salient features of feature maps. The number of feature maps of a convolution neural network will increase as its depth increases~\cite{WEBSITE:1}, this will lead to an increased number of parameters which will increase the training time. This problem can be solved using a $1\times1$ convolution layer that will do channel-wise pooling, called projection or max pooling. This technique could be used for network pruning in the CNN networks~\cite{WEBSITE:1,WEBSITE:13} and to increase the number of features after classical pooling layers. The $1\times1$ convolution layer can be used in the following three ways:
\begin{itemize}
  \item Linear projection of feature maps can be created.
  \item Since the $1\times1$ layer also works as channel-wise pooling, it can also be used for network pruning.
  \item The projection created by $1\times1$ layer can also be used to increase the number of feature maps.
\end{itemize}

\subsubsection{Downsampling with $1\times1$ filter}

A $1\times1$ filter will only have a single parameter or weight for each channel in the input that leads to single channel output value. The filter acts as a single neuron with input from the same position for each of the feature maps. The filter can be applied from left to right and top to bottom which results in a feature map of same height and width as the input~\cite{lin2013network}.

\begin{figure}
\begin{center}
\includegraphics[scale=0.35]{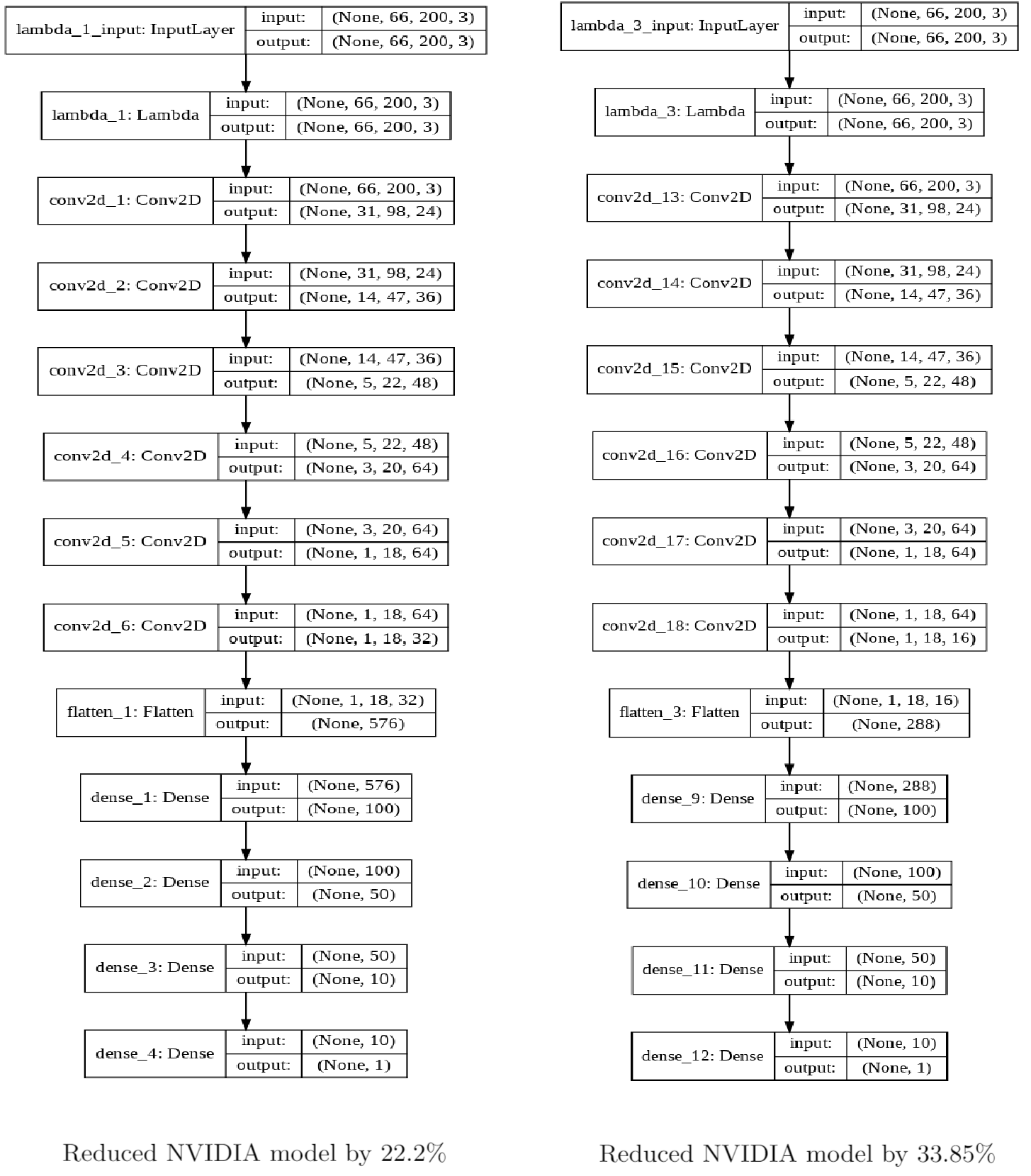}
\caption{Pruned NVIDIA architectures by using $1\times1$ filter.}
\label{fig4}       
\end{center}
\end{figure}

The idea of using of $1\times1$ filter to summarize the input feature maps is inspired from inception network proposed by Google~\cite{szegedy2015going}. The use of $1\times1$ filter allows for the effective control of number of feature maps in the resulting output. Hence, the filter can be used anywhere in the network to control the number of feature maps and so it is also called a channel pooling layer. In the two models, shown in Fig.~\ref{fig4}, the network size is pruned by 22.2\% and 33.85\% with the help of downsampling.

\subsection{Transfer learning}

Training of deep neural network needs massive computational resources. To minimize this effort, transfer learning has been explored, which assists in using neural networks implemented by various large companies who have abundant resources. The trained models provided by them can be used for academic research projects and startups~\cite{WEBSITE:2}.

\begin{figure}
\centering
\includegraphics[scale=0.35]{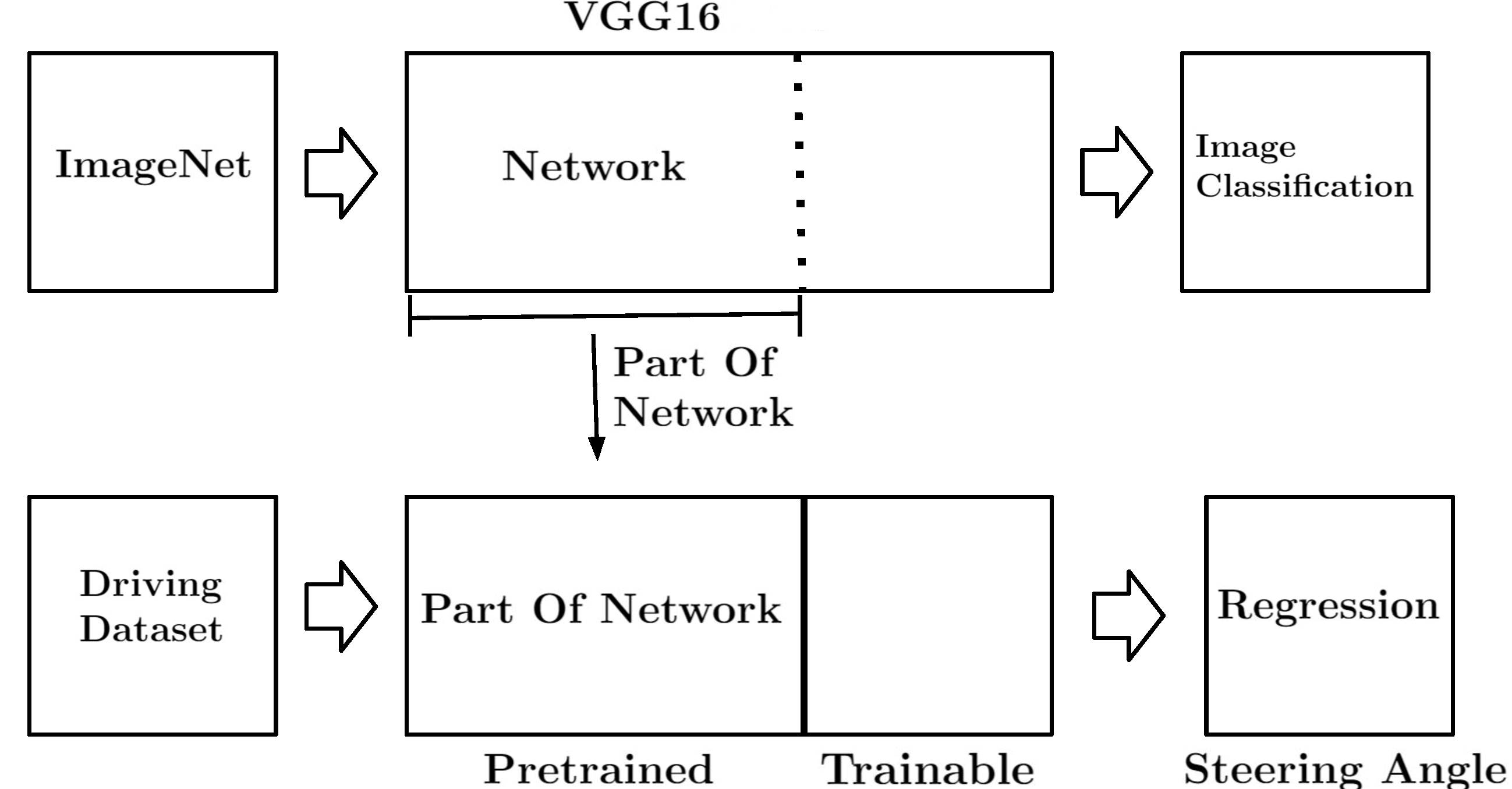}
\caption{Transfer Learning with VGG16.}
\label{fig6}       % Give a unique label
\end{figure}

As reported in recent publications the significance of the use of transfer learning for image recognition, object detection and classification, etc.~\cite{punn2020inception,punn2019crowd,punn2020monitoring} has been highlighted. In transfer learning approach a pre-trained model is adopted and fine-tuned to solve the desired problem i.e by freezing some layers and training only a few layers. Studies show that models trained on huge datasets like imagenet should generally work well for other image recognition problems~\cite{WEBSITE:10}. It is also proven in research that initializing a model with the pre-trained model weights would lead to faster convergence than initializing the model with random weights ~\cite{yosinski2014transferable}. For implementing transfer learning mechanism VGG16 has been used and all the blocks are frozen from training except the last block which contains a max-pooling layer and 3 convolution layers as highlighted in Fig.~\ref{fig6} .

Deep neural networks when trained on a huge set of images, the initial layer weights are similar regardless of the undertaken objective, whereas the end layers generally learn more problem-specific features. The initial layers of CNN learn the hidden edges, patterns and textures~\cite{WEBSITE:13} that tend to identify the features which can be utilized as generic feature extractors for identifying the desired patterns to aid in analysing the complex environment for developing an intelligent driver-less system.

\subsubsection{VGG16 with transfer learning}
VGG16~\cite{simonyan2014very} is the state-of-the-art deep CNN model which is a runner up in ILSVRC (Imagenet) competition in 2014~\cite{ILSVRC}. Compared to other models proposed in ILSVRC like ResNet50~\cite{he2016deep}, Inception\cite{szegedy2015going}, etc., VGG16 model has lesser number of parameters because of the way the convolution filters are arranged i.e $3\times3$ filter with a stride 1, followed by $2\times2$ max pool filter with stride 2. This arrangement of convolution accompanied by the max pool, is followed in the entire network consistently whereas the two fully connected layers forms the decision layer which aggregate to 138 million parameters. In the proposed approach, the initial 4 convolution blocks of the VGG16 are frozen and the last convolution block is fine tuned i.e block 5, to predict the appropriate steering angle based on the surrounding conditions acquired from the captured frames.

\section{Dataset description and preprocessing}
The dataset is a sequence of front camera dashboard view images captured around Rancho Palos Verdes and San Pedro California traffic ~\cite{DATASET:4}. It contains 45400 images and associated steering angle as described in Table~\ref{tab:1}. In this research, 80\% of images are used for training and remaining 20\% for validation testing.

\begin{table}
\centering
\caption{Dataset description.}
\label{tab:1}       
\begin{tabular}{|p{.9in}|p{3in}|}
\hline
\textbf{Feature}        & \textbf{Information}                                                      \\ \hline
Image          & The path of the image present on the disk.                       \\ \hline
Steering Angle & A value in the range of -90 to +90 indicating the steering angle. \\ \hline
\end{tabular}
\end{table}

\begin{figure}
\centering
\includegraphics[scale=0.4]{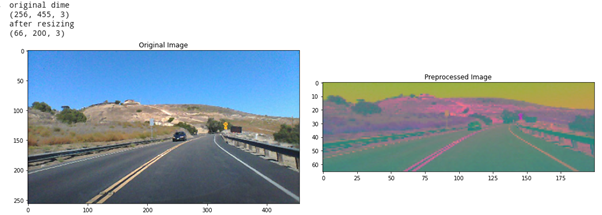}
\caption{Snapshot of original and processed images.}
\label{fig7}       
\end{figure}

The range of steering angle is between -90 to +90 where +90 indicates that the steering is tilted towards the right and -90 indicates that the steering is tilted towards the left. The data is preprocessed to get the images in the desired format which will be suitable for the network to learn and help in prediction of appropriate steering angle. The original and preprocessed images are shown in Fig.~\ref{fig7}. The images are preprocessed by performing following steps:
\begin{itemize}
  \item Remove unnecessary features by cropping the image. 
  \item Convert the image to YUV format.
  \item Reduce dimensions of the image to $66\times200\times3$.
\end{itemize}

\section{Experimental results}
Series of experiments have been carried out with baseline NVIDIA model, its pruned variants and proposed approach as described follows:
\begin{enumerate}
 \item By decreasing the number of feature maps from 64 to 32 and 64 to 16 by keeping the height and width constant, we pruned the network by 22.2\% and 33.85\% respectively.
 \item The transfer learning approach adopted with the convolution blocks of VGG16 and trained only the last block (3 convolution layers and 1 Maxpooling 2D layer).
\end{enumerate} 

The training of the models is assisted with stochastic gradient descent (SGD)~\cite{bubeck2014convex} approach with Adam as the learning rate optimizer~\cite{kingma2014adam}. For robust training, 4-fold validation technique is applied along with the earlystopping to avoid the problem of overfitting~\cite{caruana2001overfitting}. The performance of the models are evaluated using the mean squared error (MSE) as given in the Eq.~\ref{eq1}.
\begin{equation}
MSE = {\frac{1}{n} \sum_{i=1}^{n}(y_{i} - x_{i})^{2}}
\label{eq1}
\end{equation}
where $y_i$ stands for the actual steering value and $x_i$ stands for the predicted steering angle. Here, the lesser MSE indicates higher learning ability whereas a higher MSE means the model is not learning from complex environments. 

The experimental results show that the pruned networks do not perform better compared to the baseline model. It is also observed that the proposed VGG16 model with transfer learning (training only last 4 convolutions layers) is trained within 40 epochs compared to other models which are trained with 100 epochs. With the experimental results, it has been proved that the VGG16 model with transfer learning works better as compared to the other NVIDIA models. As observed from the Table.~\ref{tab2}, the novel transfer learning based approach achieved the better MSE score as compared to NVIDIA and its pruned variants. Due to the deep nature of VGG16, the architecture is able to learn complex patterns where as shallowness of NVIDIA models restrict their ability to adopt such complex environment conditions.

\begin{table}
\centering
\caption{Performance comparison of learning models.}
\label{tab2}
\resizebox{\textwidth}{!}{
\begin{tabular}{|p{0.4in}|p{3.0in}|p{0.6in}|p{0.8in}|}
\hline
\textbf{S.No.} & \textbf{Model}                                                            & \textbf{MSE}      & \textbf{Trainable Parameters} \\ \hline
1 & NVIDIA Model                                                     & 29.24848 & 252,219              \\ \hline
2 & NVIDIA model pruned by 22.2\% with  $1\times1$ filter & 41.61325 & 196,699              \\ \hline
3 & NVIDIA model pruned by 33.8\% with  $1\times1$ filter & 38.67840 & 166,859              \\ \hline
4 & VGG16 with Transfer Learning                                     & 23.97599 & 10,373,505           \\ \hline
\end{tabular}}
\end{table}

\begin{figure}
\centering
\includegraphics[scale=0.2]{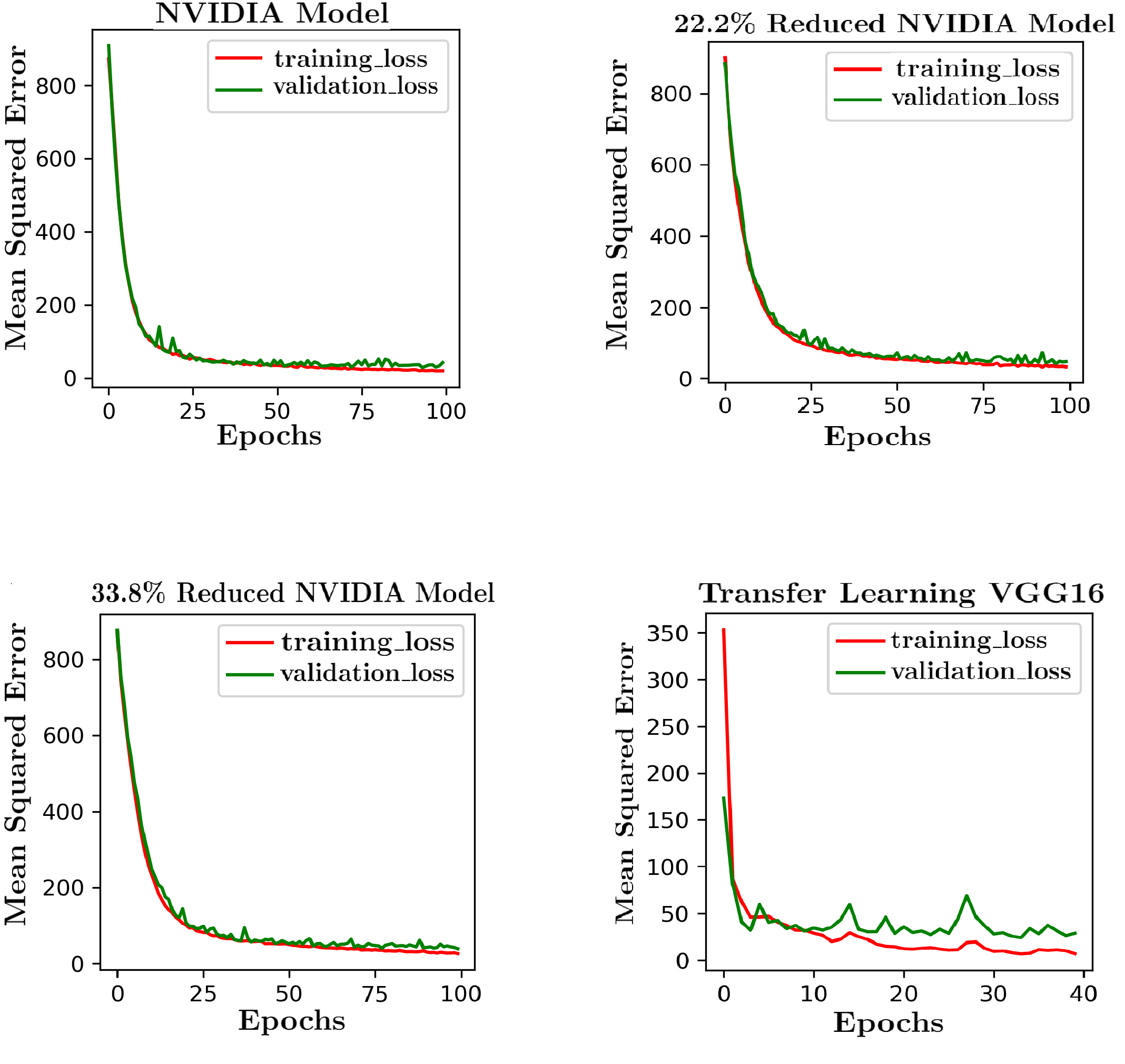}
\caption{Validation-loss Vs epochs of all the models.}
\label{fig8} 
\end{figure}

Fig.~\ref{fig8} highlights the training behavior of the models at each iteration where it is observed that the proposed approach achieved comparatively minimal loss with least number of training epochs. It is also observed that due to the adoption of trained weights the model starts with better loss and converges faster.

\section{Conclusion}
A novel approach based on transfer learning with VGG16 is proposed which is fine tuned by retraining the last block while keeping all the other layers non-trainable. The proposed model is compared with NVIDIA and its pruned architectures developed by applying $1\times1$ filter. Since the proposed transfer learning architecture starts with minimal initial loss and converges at just 40 epochs compared to NVIDIA’s architecture which took 100 epochs, experimental results show that the transfer learning based approach works better than NVIDIA and its pruned variants. Naturally, the driving patterns also depend on several other environmental conditions like weather, visibility, etc. To adopt these challenging conditions in presence of limited number of samples, generative adversarial network (GAN) can be explored in future to generate vivid weather conditions for more robust driver-less solutions.

%\begin{acknowledgements}
%If you'd like to thank anyone, place your comments here
%and remove the percent signs.
%\end{acknowledgements}

% Authors must disclose all relationships or interests that 
% could have direct or potential influence or impart bias on 
% the work: 
%
% \section*{Conflict of interest}
%
% The authors declare that they have no conflict of interest.

% BibTeX users please use one of
% \bibliographystyle{spbasic}      % basic style, author-year citations
% \bibliographystyle{spmpsci}      % mathematics and physical sciences
\bibliographystyle{spphys}       % APS-like style for physics
\bibliography{bibliography.bib}   % name your BibTeX data base

\end{document}